\RequirePackage{amsmath}
\documentclass[runningheads]{llncs}
\usepackage[T1]{fontenc}
\usepackage{graphicx,verbatim}
\usepackage{booktabs}
\usepackage{multirow}
\usepackage{placeins}
\usepackage{amssymb} 
\newcommand{\tabref}[1]{Table~\ref{#1}}
\newcommand{\figref}[1]{Fig.~\ref{#1}}
\begin{document}
%
\title{X-RAFT: Cross-Modal Non-Rigid Registration of Blue and White Light Neurosurgical Hyperspectral Images}
\titlerunning{X-RAFT: Cross-Modal Non-Rigid Registration of Blue and White Images}
%

\author{
Charlie Budd\inst{1}\and
Silvère Ségaud \inst{1} \and
Matthew Elliot\inst{1,2}\and
Graeme Stasiuk\inst{3} \and
Yijing Xie\inst{1} \and
Jonathan Shapey\inst{1,2} \and
Tom Vercauteren\inst{1}
}
\authorrunning{C. Budd et al.}
\institute{
Department of Surgical \& Interventional Engineering, School of Biomedical Engineering \&  Imaging Sciences, King's College London, London SE1 7EH\\
\email{charles.budd@kcl.ac.uk} \and
Department of Neurosurgery, King's College London Hospital NHS Foundation Trust, London SE5 9RS, UK. \and
Department of Imaging Chemistry \& Biology, School of Biomedical Engineering \& Imaging Sciences, King's College London, London SE1 7EH, UK.
}

    
\maketitle
\begin{abstract}
Integration of hyperspectral imaging into fluorescence-guided neurosurgery has the potential to improve surgical decision making by providing quantitative fluorescence measurements in real-time.
Quantitative fluorescence requires paired spectral data in fluorescence (blue light) and reflectance (white light) mode.
Blue and white image acquisition needs to be performed sequentially in a potentially dynamic surgical environment.
A key component to the fluorescence quantification process is therefore the ability to find dense cross-modal image correspondences between two hyperspectral images taken under these drastically different lighting conditions. 
We address this challenge with the introduction of X-RAFT, a Recurrent All-Pairs Field Transforms (RAFT) optical flow model modified for cross-modal inputs.
We propose using distinct image encoders for each modality pair, and fine-tune these in a self-supervised manner using flow-cycle-consistency on our neurosurgical hyperspectral data.
We show an error reduction of 36.6\% across our evaluation metrics when comparing to a naive baseline and 27.83\% reduction compared to an existing cross-modal optical flow method (CrossRAFT).
Our code and models will be made publicly available after the review process.
\end{abstract}
\section{Introduction}
\begin{figure}[tb]
    \centering
    \includegraphics[width=0.8\linewidth]{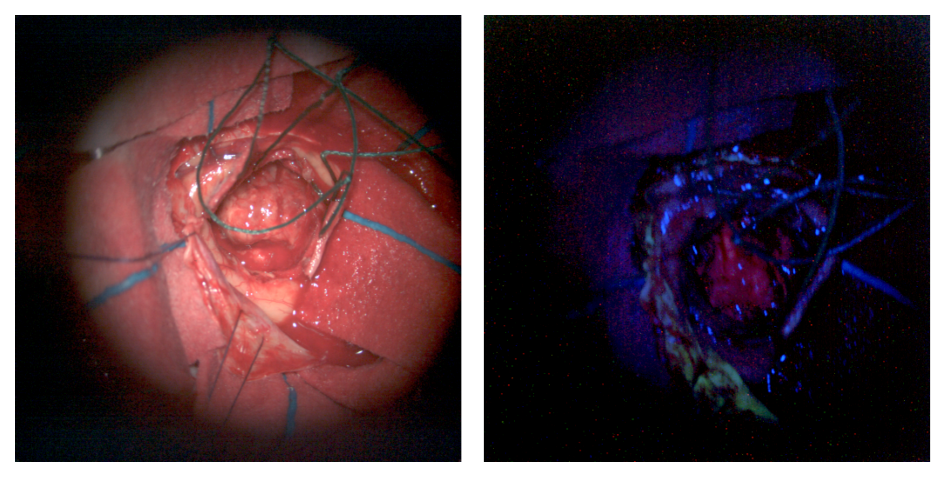}
    \caption{Exemplar white-blue image pair from our neurosurgical dataset showing a high-grade glioma visible through a craniotomy surrounded by surgical patties. Both images have been converted to RGB images for visualisation. The blue light image reveals a high level of pink fluorescence from the glioma, as-well as naturally occurring green auto-fluorescence from other tissues.
    This image pair demonstrates both homographic motion due to movement of the surgical microscope and minor non-homographic motion due to the motion of the surgical patty threads.
    }
    \label{fig:image-pair}
\end{figure}

Fluorescence guided surgery (FGS) is often employed in situations where tissue differentiation is both difficult and critical to patient outcomes.
The efficacy of FGS for glioma surgery has led to its wide adoption across neurosurgery units~\cite{elliot2025fluorescence}.
A fluorescence-inducing drug (5-ALA) is administered prior to surgery, leading to a build-up of fluorophore (PpIX) in tumour tissue which then fluoresces pink under blue light~\cite{schupper2021fluorescence}.
Neurosurgical 5-ALA-PpIX FGS is not without limitations as
surgeons still make qualitative judgment based on visual inspection of the margins of fluorescence glow, as illustrated in \figref{fig:image-pair}.
Integration of hyperspectral imaging (HSI) into 5-ALA-PpIX FGS has the potential to provide 
real-time quantitative fluorescence measurements at every pixel across the field of view~\cite{kotwal2025hyperspectral}.
However, 
accurate fluorescence
quantification in turbid biological tissue requires compensating for the varying optical characteristics of tissue~\cite{walke2023challenges}.
Combining paired spectral data from blue light fluorescence and white-light reflectance has been shown to allow for such compensation~\cite{xie2017wide}.
Due to the dynamic nature of surgery
combined with the fact that white light reflectance and blue light PpIX fluorescence HSI  cannot be acquired simultaneously, it cannot be guaranteed that any pair of white and blue light HSI data is spatially well aligned.
The white light image must therefore be spatially co-registered with the blue light image, requiring dense pixel-wise spatial correspondences.
In other words, cross-modal optical flow is needed to achieve quantitative FGS in neurosurgery.

Deep learning methods dominate the field of optical flow estimation.
Recurrent All-Pairs Field Transforms (RAFT)~\cite{teed2020raft} represents the state-of-the-art approach for it and it
demonstrates solid generalisability to 
surgical images
\cite{budd2024transferring,gerats2024neuralfields3dtracking,liu2025motionboundary}.
Optical flow methods rely on local similarity between corresponding image regions.
This presents a challenge when working with cross-modal image pairs, as the image regions to be matched 
cannot trivially be compared.
While some recent works have focused on cross-modal homography estimation~\cite{jiang2023breaking,li2024towards}, this is not sufficient as our image pairs feature a significant amount of non-homographic motion.
Application of multimodal registration designed for medical imaging data~\cite{chen2025survey} doesn't translate easily to surgical images due to large tissue motion and the projective nature of surgical microcopy views.
Closer to our task, CrossRAFT~\cite{zhou2022promoting} and CrossMCMRAFT~\cite{zhai2023cross}, both attempt to train optical flow models to be robust to an open-set of modalities by training with single modality image pairs with differing random augmentations applied.
This approach is attractive for a number of reasons, namely that image pairs from a single modality may be used for training, and that the resulting model would in theory generalise to any pair of modalities.
However, this generalisability is heavily dependant on the augmentations employed.

In this work, we propose X-RAFT, the first learning-based cross-modal optical flow approach in a surgical setting.
Our main novel contribution, lies in leveraging distinct modality-specific image encoders within a modified RAFT architecture.
This is in contrast to CrossRAFT and CrossMCMRAFT which both take the approach of attempting to unify the encoded latent representations.
Another main contribution, key to enabling our objectives in the abscence of large-scale ground-truth data, is the introduction of
a self-supervised training approach that requires only unaligned image triplets of mixed modality.
Our training approach allows
X-RAFT to learn intricacies of the modalities 
beyond what domain randomisation can achieve.
Our experiments show that X-RAFT provides significant performance increase over our baselines.

\section{Material and Methods}

\subsubsection{Neurosurgical dataset}
Our data consists of 52 hyperspectral videos of 
glioma surgeries.
The videos were captured with a 10 band hyperspectral camera integrated into a surgical microscope as part of an ethically-approved clinical study\footnote{Ethics information  redacted for the anonymous review process}.
Each video consist of sections recorded under white light illumination followed by blue light illumination used by the surgeon for fluorescence visualisation.
Due to the low quantum efficiency of PpIX, exposure times in the range of 1.5 to 2 seconds are used to capture blue light HSI.
Even then, the blue images 
present
a lower signal-to-noise-ratio than the white counterparts, as
shown in~\figref{fig:image-pair} where colour imaging is reconstractued from high-dimensional HSI data~\cite{Li_2024_BMVC}.

While 
surgeons were instructed 
to target static scenes during the recording, 
the reality of recording in a high-stakes environment meant that around 60\% of the data contains a noticeable level of motion.
This motion arises from a number of factors, 
including
handling of surgical tools,
manipulation of brain tissue, 
pulsatility of the patient brain,
and sometimes repositioning of the surgical microscope.
Through visual inspection, we manually split our dataset into 32 videos with clearly visible motion and 20 videos with little-to-no motion.
We manually selected around 10-20 white and blue images per video based on various qualitative assessments such as lack of motion blur and if the image is well exposed.
We then annotated our high motion data to aid in evaluation.
Specifically, a trained neurosurgeon selected a white+blue image pair from each high motion video, forming a reserved testing split.
They then created binary segmentations of visible tumour tissue and marked several corresponding key point pairs of various elements in both the white and the blue frame.

\subsubsection{Baseline approaches}
We choose RAFT as a baseline due to its availability, popularity, and generalisable performance to surgical scenarios.
As RAFT is intended for RGB images, not hyperspectral data, we first convert our white light and blue light HSI data to RGB images by applying a colour space conversion matrix.
Different colour spaces can be targeted for different purposes, with standard RGB (sRGB) allowing intuitive human vision like visualisation~\cite{Li_2024_BMVC}.
A naive approach to bridge the domain gap between our blue and white images
is to apply different handcrafted colour transformations to each image to increase the pixel-wise similarity of the RGB reconstruction.
There are two main contributors of signal in our blue light HSI data: pink and green fluorescence and residual blue reflectance not rejected by the excitation light rejection filter.
Conversely, our white light HSI contain only reflectance data but from across the visible spectral range.
As such, only blue reflectance data is common signal across the two modalities.
We therefore choose to take the blue channel from both reconstructed sRGB images, repeating this three times to construct greyscale (BBB) images for inputting into RAFT.
We present results for RAFT running on both sRGB and BBB images in the results section.
We additionally make use of CrossRAFT~\cite{zhou2022promoting} as a additional baseline specifically aimed at cross-modal inference.
We tried both the pretrained weights provided as well as running their published training pipeline on our neurosurgical data.
The former produced the best results in all metrics by a fair margin and so we report only these results for brevity.
Unfortunately CrossMCMRAFT~\cite{zhai2023cross} cannot be compared as there is no implementation or model released with this work.

\subsubsection{Our proposed X-RAFT}
\begin{figure}[tb]
    \centering
    \includegraphics[width=0.9\linewidth]{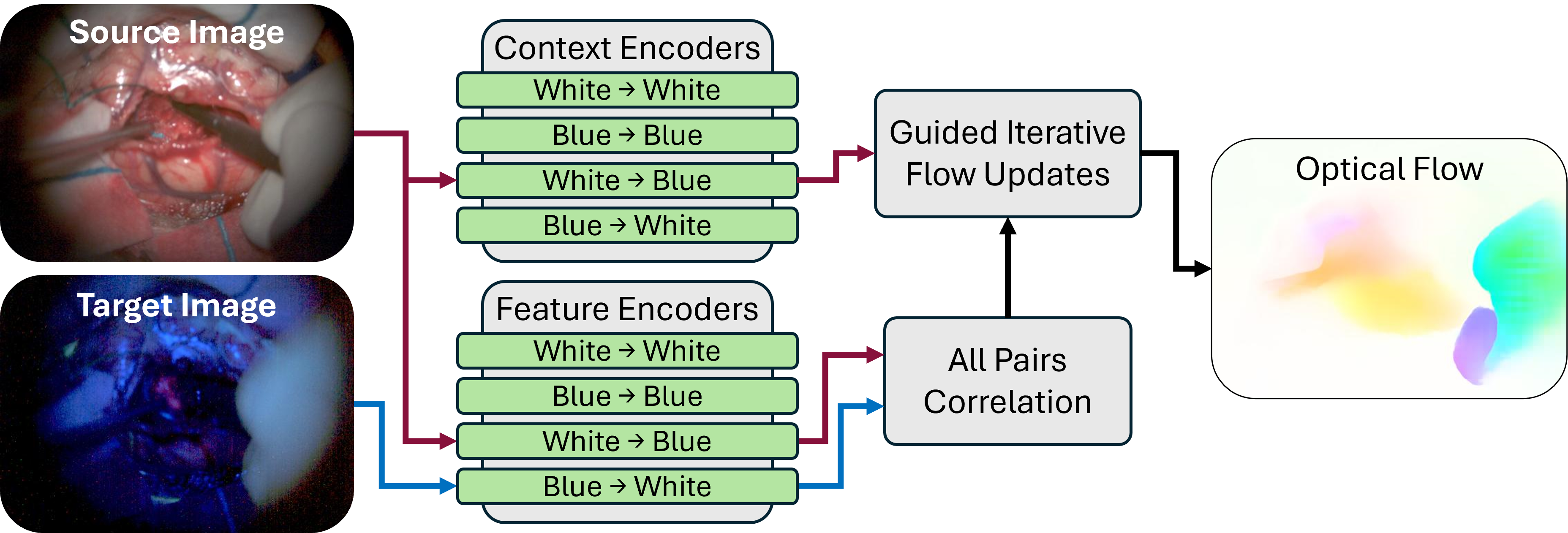}
    \caption{Model architecture of X-RAFT. Our modifications to the original RAFT architecture simply add duplicate feature and context encoders for each modality pair.}
    \label{fig:model}
\end{figure}

RAFT estimates optical flow through a three-stage process.
Firstly, the source and target images are independently passed through a feature encoder to produce latent feature embeddings with reduced spatial resolution.
These embeddings are then used to construct a dense all-pairs correlation volume, capturing potential correspondences between every location in the two images.
Finally, the optical flow map is generated through a series of iterative updates guided by a separate encoding of the source image, produced by a context encoder.
Due to the significant difference in local image features between our white and blue images, we hypothesise that it will be beneficial to learn different feature and context encoders for these modalities.
We extend this thinking further by learning a distinct feature and context encoder for each possible modality pair (e.g. blue-to-blue, blue-to-white, white-to-blue, and white-to-white).
In this way, each encoder can focus on extracting features that will best help find correspondences in the other modality.
As an example, to infer optical flow from a white source image to a blue target image, as depicted in~\figref{fig:model}, the white source image is encoded by the white-to-blue feature and context encoders, while the blue target image is encoded by the blue-to-white feature encoder.
Note that we do not make this directional, i.e. the white-to-blue encoder is used to encode white images whether they are the source or the target image, so long as the other image in the pair is a blue image.

\paragraph{Scenario-specific choices}
For our naive baseline method, we reasoned that the blue channel is of most importance to finding correspondences.
We would like to provide this prior information to the model whilst still allowing the model to learn helpful features which may be extracted from the other channels.
To achieve this, we carefully 
initialise
the weights of the first convolutional layer for all cross-modal encoders to produce the same output for an RGB image as if we had instead provided a BBB image.
In the resulting X-RAFT RGB,
the weights $B$ are zeroed for the red and green channels, while the blue channel is assigned the channel-wise sum of the pre-trained weights $A$:
\begin{equation}
B_{niwh} = 
\begin{cases}
\sum_{i} A_{niwh}, & \text{if } i = 3 \\
\mathbf{0}, & \text{otherwise} \\
\end{cases}
\end{equation}

As our 10 band HSI data contains information beyond our RGB reconstruction, we also propose X-RAFT HSI which accepts 10 input channels.
We again modify the first layer of each image encoder.
The weights of this layer $C$ are efficiently initialised by multiplying the existing weights $B$ by the colour space conversion matrix $Q$, used to convert the hyperspectral images into RGB images.
\begin{equation}
C_{n c w h} = \sum_i B_{n i w h} \cdot Q_{i c}
\end{equation}
This provides the same results as running on RGB images but removes the 3 channel bottleneck, thus allowing the model to learn to use the new channel information without degrading initial performance.

\begin{figure}[tb]
    \centering
    \includegraphics[width=0.8\linewidth]{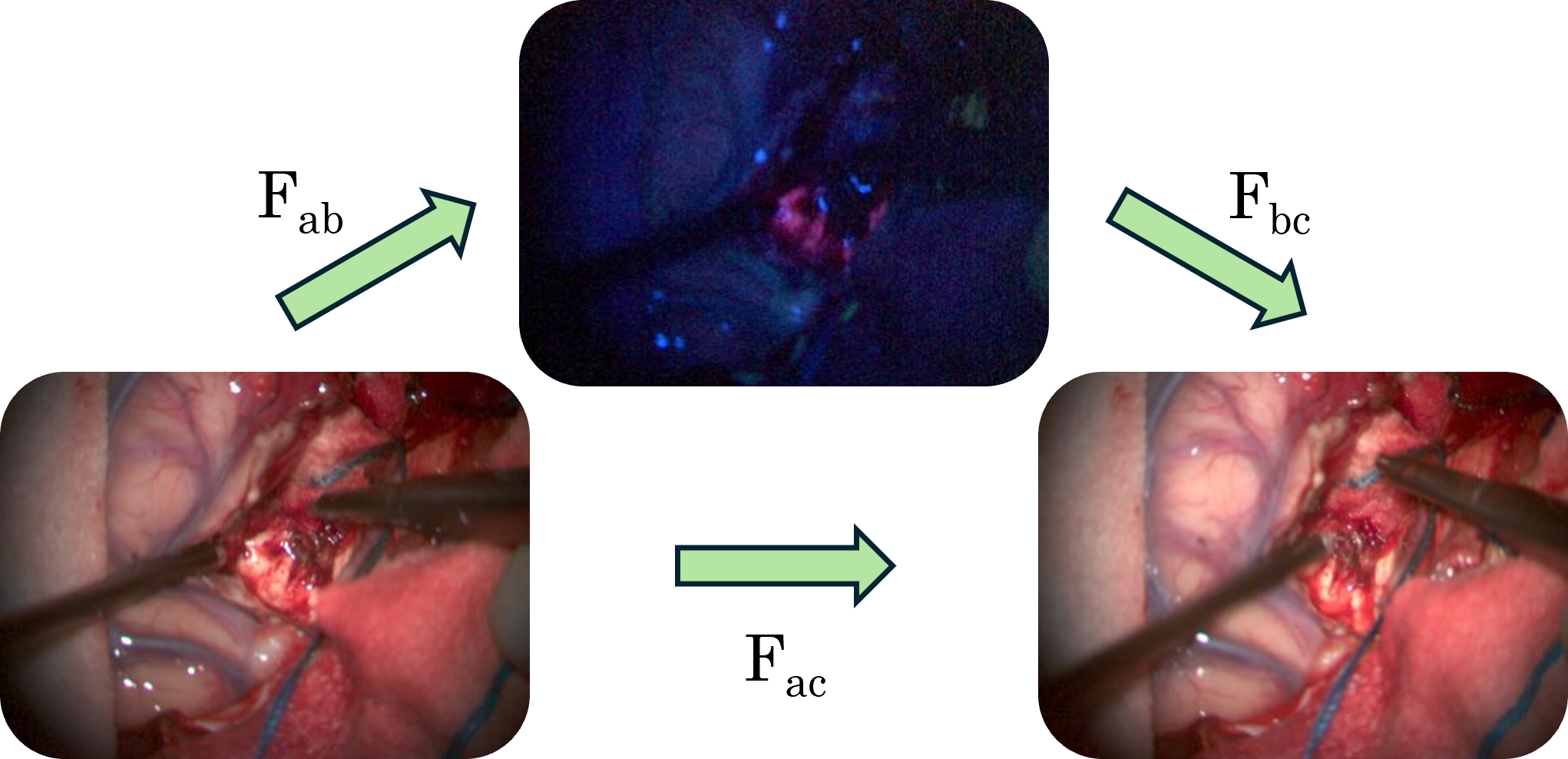}
    \caption{Depiction of the flow cycle-consistency triplet for our self-supervision X-RAFT training. Optical flow inferred between the two white images, using off-the-shelf RAFT, is used as supervision for training of the white-to-blue and blue-to-white flow.}
    \label{fig:flow-cycle}
\end{figure}

\paragraph{Flow-cycle-consistency loss}
As we lack ground truth optical flow for our data, we require a self-supervised training strategy.
Typical methods in optical flow rely on either flow consistency or photometric constancy.
In our case, photometric constancy is inappropriate as the image pairs would not be expected to be visually similar after registration.
As such, we choose to design a flow consistency based learning objective.
We take advantage of the fact that we posses multiple white light images of each scene, between which we may infer reliable optical flow using off-the-shelf pre-trained RAFT.
We therefore construct a flow cycle around a triplet constituting of two different white images and a blue image sampled from the same video, depicted in~\figref{fig:flow-cycle}.
Our loss function may then be expressed as the Euclidean distance between the vectors in the two optical flow fields, known as the end point error (EPE):
\begin{equation}
    L=\text{EPE}(F_{ab} + \text{warp}(F_{bc}, F_{ab}), F_{ac})
\end{equation}
where $F_{ij}$ represents the optical flow from the image $I_i$ to the image $I_j$, and $\text{warp}(A, F)$ pulls pixels in the entity $A$ backwards along the flow field $F$ through grid sampling.
Spatial correspondences between two images may not exist everywhere or information may be missing to extract them meaningfully.
Our training objective is thus adapted by attempting to mask out such regions.
One main consideration is occlusion induced by motion between frames.
A typical approach for determining occluded regions is to calculate the optical flow in both directions.
If true correspondences are found, the flow fields should have low discrepancy:
\begin{equation}
    M^o_{ij} = (F_{ij} - \text{warp}(F_{ji}, F_{ij})) > \varepsilon^o
\end{equation}
Additionally we note that due to the low signal in much of the blue image, a significant part of the image can be within the noise floor (i.e. black).
We define a second mask by thresholding the average pixel intensity across all channels:
\begin{equation}
    M^d_{i} = \frac{1}{C} \sum_{c=1}^C I_{i,c} > \varepsilon^d
\end{equation}
For consistency, the channel average is computed from HSI, even if the image is then converted into RGB for inputting to the model.
This determines where there is little to no feature detail to use in finding correspondences.
A final mask relative to $I_a$ may then be calculated as the combination (pixel-wise multiplication) of these masks, warped where appropriate to align with the source image:
\begin{equation}
    M = M^o_{ac} \cdot M^o_{ab} \cdot \text{warp}(M^o_{bc}M^d_{b}, F_{ab}).
\end{equation}
This mask is then used to mask our loss function so that we only provide supervision for image regions we believe to have findable correspondences.

\paragraph{Model training}
To avoid overfitting to our approximate training objective and small dataset, we choose to freeze all parameters of the model, leaving only the cross-modal feature and context encoders unfrozen.
We iterate through our high motion training data sampling batches of white-blue-white triplets and optimising our flow-cycle-consistency loss for the first three flow field iterations.
Mask thresholds of $8.0$ and $0.07$ were used for $\varepsilon^o$ and $\varepsilon^d$ respectively.
We used a batch size of $20$ and an Adam optimiser with a learning rate of $5\times10^{-5}$.
We validate against our synthetic flow metric, as described in the Evaluation Section, every $10$ batches and early stop with a patience of $20$, generally our models trained for around 450 batches taking about an hour on a single 40Gb A100 GPU.
All hyper parameters were decided on via a light and informally hyper parameter tuning based on the validation performance.

\subsubsection{Evaluation}
Due to the lack of ground truth pixel-wise correspondences, 
we choose to employ a mixed evaluation strategy using approximate, sparse, or secondary metrics to ensure robust method ranking.
Each metric is evaluated in both inference directions, white to blue and blue to white, as well as the combination of the two.
Firstly, we generate synthetic flow fields which are then used to apply elastic deformation to one image from a white-blue pair sampled from our low motion dataset.
We make the assumption that the only motion between the image pair is that which is induced by the deformation, thus the generated optical flow field provides pseudo ground truth.
Following RAFT, we use endpoint error (EPE) to measure the accuracy for the predicted optical flow.
While this provides dense error measurement, the generated motion is not necessarily realistic and may not account for all motion between the images.
To address this we also make use of our annotated image pairs from our high motion dataset.
The annotated key point pairs are used to provide sparse but precise evaluation of the inferred flow fields, again using EPE.
Finally, the inferred optical flow fields are then used to propagate the annotated tumour segmentation masks between the two images.
We take 1$-$IoU of the annotated and propagated mask as an error metric for consistency with the other error metrics.

\section{Results and discussion}
\begin{figure}[tb]
    \centering
    \includegraphics[width=\linewidth]{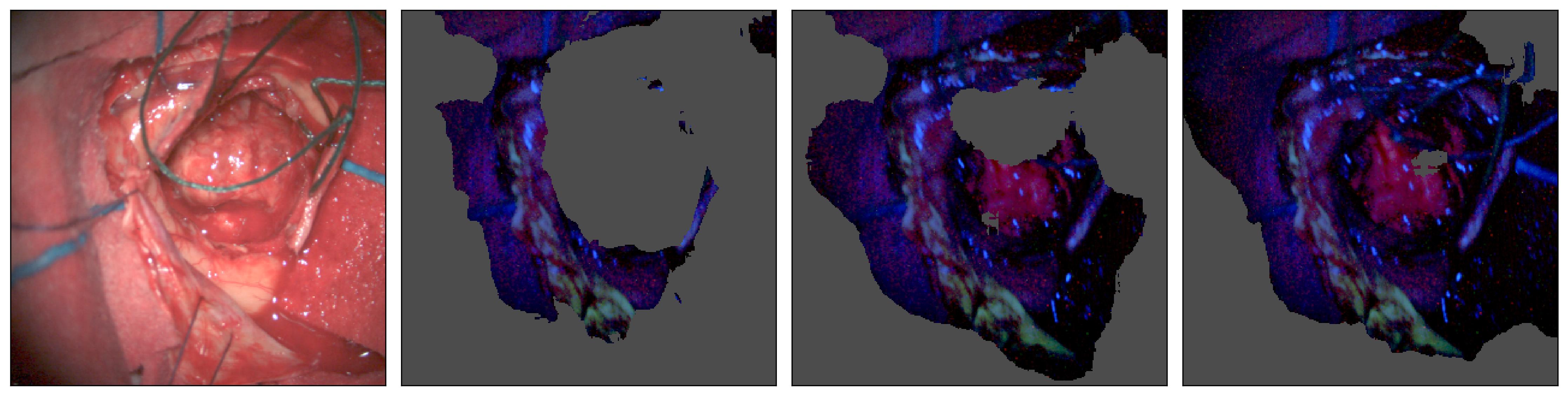}
    \caption{An example registration of a blue image onto a white image using, from left to right, RAFT BBB, CrossRAFT, and X-RAFT HSI. Flow discrepancy with a threshold of 3 pixels is used to discard uncertain correspondences (shown in gray).}
    \label{registration}
\end{figure}

We present evaluation metrics for all methods in \tabref{results}.
This shows that X-RAFT running on hyperspectral images provides an average error reduction across the three metrics of 36.6\% compared to RAFT on our transformed BBB images and 27.8\% compared to CrossRAFT on RGB images.
Discarding the synthetic flow result from the calculation of percentage improvements, for which CrossRAFT performs uncharacteristically poorly, we still find a 9.0\% improvement.
In addition we present an exemplar blue-white registration in \figref{registration} demonstrating an increase in found correspondences during registration. 
\begin{table}
\centering
\caption{Evaluation metrics for different optical flow models presented as means and standard deviations over 5 training runs where appropriate. Scores are reported separately for white-to-blue and blue-to-white inference directions as well as the average.}
\label{results}
\setlength{\tabcolsep}{2pt}
\begin{tabular}{c|c|c|c|c}
\toprule
Model & Direction & Synthetic (EPE $\downarrow$) & Keypoint (EPE $\downarrow$) & Mask ($1{-}$IoU $\downarrow$) \\
\midrule
\multirow{3}{*}{RAFT RGB}
  & W$\rightarrow$B  & 26.636              & 6.050               & 0.260 \\
  & B$\rightarrow$W  & 19.617              & 12.980              & 0.373 \\
  & Both             & 23.127              & 9.515               & 0.316 \\
\midrule
\multirow{3}{*}{RAFT BBB}
  & W$\rightarrow$B  & 8.084               & 6.149               & 0.230 \\
  & B$\rightarrow$W  & 6.815               & 10.972              & 0.364 \\
  & Both             & 7.449               & 8.561               & 0.297 \\
\midrule
\multirow{3}{*}{CrossRAFT RGB}
  & W$\rightarrow$B  & 11.475  & 6.411  & 0.200 \\
  & B$\rightarrow$W  & 16.323  & 6.241  & 0.186 \\
  & Both             & 13.899  & 6.313  & 0.193 \\
\midrule
\multirow{3}{*}{X-RAFT RGB}
  & W$\rightarrow$B  & 5.183$\pm$.107      & 5.649$\pm$.226      & \bf{0.181$\pm$.004} \\
  & B$\rightarrow$W  & 4.484$\pm$.021      & 5.719$\pm$.073      & \bf{0.184$\pm$.005} \\
  & Both             & 4.834$\pm$.046      & 5.684$\pm$.137      & \bf{0.182$\pm$.004} \\
\midrule
\multirow{3}{*}{X-RAFT HSI}
  & W$\rightarrow$B  & \bf{5.112$\pm$.094} & \bf{5.275$\pm$.229} & 0.184$\pm$.004 \\
  & B$\rightarrow$W  & \bf{4.466$\pm$.023} & \bf{5.608$\pm$.115} & 0.185$\pm$.002 \\
  & Both             & \bf{4.789$\pm$.047} & \bf{5.442$\pm$.129} & 0.185$\pm$.003 \\
\bottomrule
\end{tabular}
\end{table}

In this work we have identified a novel integration problem requiring the first investigation of learning-based cross-modal optical flow in a surgical setting.
The novel architectural modifications and self-supervised fine-tuning methodology of our proposed X-RAFT model demonstrates a clear improvement over the existing state-of-the-art approach.
A primary limitation of this work is the use of frames from the same videos as frames used to train the models.
This was done to ensure a significant amount of evaluation data was available.
While we aimed to prevent overfitting by validating on the low-motion data not used in training, and by only training the encoders, it is still possible that the high correlation with the training data may have introduced bias.
Our future work will see the application of this methodology to aid in the quantification of fluorescence measurement by allowing the use of the 
co-register white light images to compensate for optical properties in the blue light image space.

%

\FloatBarrier

\bibliographystyle{splncs04}
\bibliography{library}

\begin{thebibliography}{10}
\providecommand{\url}[1]{\texttt{#1}}
\providecommand{\urlprefix}{URL }
\providecommand{\doi}[1]{https://doi.org/#1}

\bibitem{budd2024transferring}
Budd, C., Vercauteren, T.: Transferring relative monocular depth to surgical vision with temporal consistency. In: International Conference on Medical Image Computing and Computer-Assisted Intervention. pp. 692--702. Springer (2024)

\bibitem{chen2025survey}
Chen, J., Liu, Y., Wei, S., Bian, Z., Subramanian, S., Carass, A., Prince, J.L., Du, Y.: A survey on deep learning in medical image registration: New technologies, uncertainty, evaluation metrics, and beyond. Medical Image Analysis  \textbf{100},  103385 (Feb 2025)

\bibitem{elliot2025fluorescence}
Elliot, M., S{\'e}gaud, S., Lavrador, J.P., Vergani, F., Bhangoo, R., Ashkan, K., Xie, Y., Stasiuk, G.J., Vercauteren, T., Shapey, J.: Fluorescence guidance in glioma surgery: A narrative review of current evidence and the drive towards objective margin differentiation. Cancers  \textbf{17}(12), ~2019 (2025)

\bibitem{gerats2024neuralfields3dtracking}
Gerats, B.G.A., Wolterink, J.M., Mol, S.P., Broeders, I.A.M.J.: Neural fields for 3d tracking of anatomy and surgical instruments in monocular laparoscopic video clips (2024), \url{https://arxiv.org/abs/2403.19265}

\bibitem{jiang2023breaking}
Jiang, Z., Zhang, Z., Liu, J., Fan, X., Liu, R.: Breaking modality disparity: Harmonized representation for infrared and visible image registration. arXiv preprint arXiv:2304.05646  (2023)

\bibitem{kotwal2025hyperspectral}
Kotwal, A., Saragadam, V., Bernstock, J.D., Sandoval, A., Veeraraghavan, A., Vald{\'e}s, P.A.: Hyperspectral imaging in neurosurgery: a review of systems, computational methods, and clinical applications. Journal of Biomedical Optics  \textbf{30}(2),  023512--023512 (2025)

\bibitem{li2024towards}
Li, H., Dong, S., Wang, J., Fu, R., Jing, M., Liang, J., Fan, H., Ji, R.: Towards rgb-nir cross-modality image registration and beyond. arXiv preprint arXiv:2405.19914  (2024)

\bibitem{Li_2024_BMVC}
Li, P., MacCormac, O., Shapey, J., Vercauteren, T.: A self-supervised and adversarial approach to hyperspectral demosaicking and rgb reconstruction in surgical imaging. In: 35th British Machine Vision Conference 2024, {BMVC} 2024, Glasgow, UK, November 25-28, 2024. BMVA (2024), \url{https://papers.bmvc2024.org/0188.pdf}

\bibitem{liu2025motionboundary}
Liu, Y., Wu, P., Huo, J., Zhang, G., Yuan, Z., Bergeles, C., Sparks, R., Dasgupta, P., Granados, A., Ourselin, S.: Motion-boundary-driven unsupervised surgical instrument segmentation in low-quality optical flow (2025), \url{https://arxiv.org/abs/2403.10039}

\bibitem{schupper2021fluorescence}
Schupper, A.J., Rao, M., Mohammadi, N., Baron, R., Lee, J.Y., Acerbi, F., Hadjipanayis, C.G.: Fluorescence-guided surgery: a review on timing and use in brain tumor surgery. Frontiers in Neurology  \textbf{12},  682151 (2021)

\bibitem{teed2020raft}
Teed, Z., Deng, J.: Raft: Recurrent all-pairs field transforms for optical flow. In: Computer Vision--ECCV 2020: 16th European Conference, Glasgow, UK, August 23--28, 2020, Proceedings, Part II 16. pp. 402--419. Springer (2020)

\bibitem{walke2023challenges}
Walke, A., Black, D., Valdes, P.A., Stummer, W., K{\"o}nig, S., Suero-Molina, E.: Challenges in, and recommendations for, hyperspectral imaging in ex vivo malignant glioma biopsy measurements. Scientific reports  \textbf{13}(1), ~3829 (2023)

\bibitem{xie2017wide}
Xie, Y., Thom, M., Ebner, M., Wykes, V., Desjardins, A., Miserocchi, A., Ourselin, S., McEvoy, A.W., Vercauteren, T.: Wide-field spectrally resolved quantitative fluorescence imaging system: toward neurosurgical guidance in glioma resection. Journal of Biomedical Optics  \textbf{22}(11),  116006--116006 (2017)

\bibitem{zhai2023cross}
Zhai, M., Ni, K., Xie, J., Gao, H.: Cross-modal optical flow estimation via modality compensation and alignment. In: ICASSP 2023-2023 IEEE International Conference on Acoustics, Speech and Signal Processing (ICASSP). pp.~1--5. IEEE (2023)

\bibitem{zhou2022promoting}
Zhou, S., Tan, W., Yan, B.: Promoting single-modal optical flow network for diverse cross-modal flow estimation. Proceedings of the AAAI Conference on Artificial Intelligence  \textbf{36}(3),  3562--3570 (Jun 2022)

\end{thebibliography}
\end{document}